\documentclass[letterpaper]{article} 
\usepackage{aaai25}  
\usepackage{times}  
\usepackage{helvet}  
\usepackage{courier}  
\usepackage[hyphens]{url}  
\usepackage{graphicx} 
\usepackage{natbib}  
\usepackage{caption} 
\usepackage{todonotes}
\usepackage{glossaries}

\usepackage{amsmath,amssymb}
\usepackage{rotating}
\usepackage{xcolor}
\usepackage{makecell}
\usepackage{multirow}
\usepackage{subcaption}

\usepackage[hyphens]{url}

\newacronym{ml}{ML}{Machine Learning}

\title{How Green Can AI Be?\\A Study of Trends in Machine Learning Environmental Impacts}
\author{
    Cl\'ement Morand\textsuperscript{\rm 1},
    Anne-Laure Ligozat\textsuperscript{\rm 1,\rm 2},
    Aur\'elie N\'ev\'eol\textsuperscript{\rm 1}
}
\affiliations {
    \textsuperscript{\rm 1}Université Paris-Saclay, CNRS, LISN,\textsuperscript{\rm 2}ENSIIE\\
    name.surname@lisn.upsaclay.fr
}

\newcommand{\KILL}[1]{}
\newcommand{\COtwo}{CO\textsubscript{2}~eq}
\newcommand{\Sbe}{Sb~eq}

\begin{document}

\maketitle

\begin{abstract}

The compute requirements associated with training \textit{Artificial Intelligence} (AI) models 
have increased exponentially over time.
Optimisation strategies aim to reduce the energy consumption and environmental impacts associated with AI, possibly shifting impacts from the use phase to the manufacturing phase in the life-cycle of hardware.
This paper investigates the evolution of individual graphics cards production impacts and of the environmental impacts associated with training \gls{ml} models over time.

We collect information on graphics cards used to train \gls{ml}  models and released between 2013 and 2023. We assess the environmental impacts associated with the production of each card to visualize the trends on the same period. Then, using information on notable AI systems from the Epoch AI dataset we assess the environmental impacts associated with training each system.

The environmental impacts of graphics cards production have increased continuously. 
The energy consumption and environmental impacts associated with training models have increased exponentially, even when considering reduction strategies such as location shifting to places with less carbon intensive electricity mixes.

These results suggest that current impact reduction strategies cannot curb the growth in the environmental impacts of AI. 
This is consistent with rebound effect, where the efficiency increases fuel the creation of even larger models thereby cancelling the potential impact reduction. Furthermore, these results highlight the importance of considering the impacts of hardware over the entire life-cycle rather than the sole usage phase in order to avoid impact shifting. 

The environmental impact of AI cannot be reduced without reducing AI activities as well as increasing efficiency.

\end{abstract}

\section{Introduction}

Environmental impacts such as the carbon footprint, water usage, metallic resource depletion or toxicity to human and non human life
caused by \gls{ml} raise increasing concern. Strategies that optimize the energy consumption of models training have been developed to mitigate the environmental impacts of the sector. \textit{Technical optimisations} adjust model architecture to offer the same task performance in smaller models. \textit{Shift optimisations} displaces computation towards less carbon-intensive mixes performing compute when more renewable energy is available or purchasing of renewable energy. 
Current strategies also rely on frequent hardware updates to benefit from more energy efficient recent hardware.

Major AI companies have claimed that these strategies would mitigate and eventually reduce the carbon footprint of \gls{ml} training \cite{Patterson2022plateau}. However, they fail to account for two important factors. First, frequent hardware upgrades cause impact shifting (i.e., reducing the environmental impacts from one life cycle phase or impact category at the detriment of other life cycle phases and/or impact categories), which limits the overall benefit. Second, optimisation often accompany rebound effect (i.e., optimization can lead to a less than expected decrease, or even increase the total environmental impacts of the sector).

Herein, we consider impact shifting, rebound effects and other digital trends to question "How green can AI be?"
Specifically, we investigate how the impacts of individual graphics cards have evolved in the past 10 years and how the environmental impacts associated with training \gls{ml} models have  evolved over this period. We focus on the past decade (2013-2023) and conduct a detailed study of the characteristics and environmental impacts of the production of NVIDIA workstation graphics cards. We link information on hardware to \gls{ml} models trained using them. 
Finally, we question the global effect of the impact mitigating strategies. 

The main contributions of this paper are as follows:
\begin{itemize}
    \item An assessment of the environmental impact of graphic cards released between 2013 and 2023
    \item A detailed analysis of the environmental impact of \gls{ml} models, including impact shifting and rebound effect
    \item Evidence that the impact of AI is increasing in spite of the promises of current reduction strategies. The material used in our analysis are shared to foster reroducibility.
\end{itemize}

\section{State of the art}
\label{sec:sota}

After the high level of carbon emissions associated with training Natural Language processing models was reported~\cite{Strubell2019energy}, researchers stated the need for a "Green AI"~\cite{Schwartz2020green}, soon structured as an entire research field~\cite{Verdecchia2023systematic}. 
Green AI research includes the development of carbon footprint reporting tools, such as {\em Green Algorithms} \cite{Lannelongue2021green} or {\em CarbonTracker} \cite{anthony2020carbontracker}. Recent reviews offer a comparison of tools focusing on measuring the impact of the use phase of computation~\cite{Bouza2023estimate,Jay2023experimental,Bannour2021evaluating}. Subsequently, {\em MLCA} was developed to also account for other phases of the life cycle of the hardware used and for abiotic depletion potential and primary energy demand in addition to the carbon footprint \cite{Morand2024MLCA}.

These tools have been used for individual reporting of the carbon footprint associated with training \gls{ml} models, e.g., \cite{Luccioni2022estimating}. Optimisation techniques have been developed and used to build less carbon intensive models \cite{Wu2022sustainable, Patterson2022plateau}.
However, we do not know of studies of the environmental impacts at the scale of the \gls{ml} sector. \citet{Thompson2023Computational} and \citet{sevilla2022compute} have shown that compute requirements for \gls{ml} models training over time follow an exponential growth. 
\citet{Desislavov2023trends} studied the energy demand for \gls{ml} models inferences over time, also showing a tendency towards a growth in energy demand per inference over time. A position paper by \citet{devries2023growing} discussed the potential growth of the \gls{ml} sector with the recent surge in demand for large language models freely accessible as chatbots.

The energy efficiency of  graphics cards has increased exponentially which leads some to think that frequently replacing hardware will shrink the carbon footprint of training models \cite{Patterson2022plateau}. However, the computation requirements to train models have also increased exponentially \cite{sevilla2022compute,Thompson2023Computational}. This seems to indicate the presence of a \emph{producer rebound}  effect where the efficiency increases fuel the creation of even larger models thereby canceling the potential impact reduction \cite{Coroama2019rebound}.
\citet{Bol2021Moore} and \citet{Gossart2015rebound} showed that optimisations are often absorbed by the growth of the \emph{Information and Communication Technologies} (ICT) sector. Does this observation also hold for the case of the AI sector?
\citet{Masanet2020recalibrating} has shown that despite an exponential growth in computation demand between 2012 and 2018, the total energy consumption of datacenters had only increased by around 5\% indicates that "A [...] source of higher electricity consumption is coming from energy-intensive data centres, artificial intelligence (AI) and cryptocurrencies, which could double by 2026."

Overall, previous work led to the development of tools for measuring the impact of specific AI algorithms. Studies at scale have addressed parts of the AI sector only such as datacenters, or have omitted rebound effect and hardware production impacts. Herein, we present a comprehensive study of the impact of \gls{ml} hardware and models over a decade, including rebound effect and hardware production.

\section{Methodology}
\label{sec:methods}

Details on the sources, processes and methodological choices are available in the accompanying code and data.

\subsection{Graphics cards production impacts}

\label{subsec:graphics-cards}

To better understand the extent of the impact shifting, we study the evolution of graphics cards characteristics over time and the associated evolution in the environmental impacts of producing graphics cards over time. Life cycle assessments of ICT equipment have shown the importance of \textit{Integrated Circuits (IC)} in the environmental impacts of ICT equipment \cite{Clement2020sources}. ICs come in two forms in graphics cards: GPU (logic type ICs) and memory (memory type ICs). The surface of the GPU is indicated by the GPU \textit{die area}. 
Contributors to the impact of producing ICs include the surface of the IC (i.e., the die area of the GPU and the surface of memory type IC for the memory chips), as well as how finely the circuits are printed on the semiconducting material. 
Thus, if the die area increases we can expect an increase in the environmental impacts of the device. Furthermore, \cite{Pirson2023environmental} have shown that, with finer technological nodes, the environmental impacts per produced cm$^2$ of die increase. Thus, as the quantity of memory increases (probable increase in memory type IC surface), the GPU die size increases and the technological node gets finer (latest GPUs processed at 5nm), we can estimate that the environmental impacts of graphics cards production increases. 

We study the evolution of the characteristics of graphics cards over time to check these hypotheses. We focus on NVIDIA workstation graphics cards as NVIDIA is the lead provider of graphics cards for servers. Additional cards frequently used to train \gls{ml} models are also included in our study: Google's TPU, Huawei or Cerebras or NVIDIA non workstation cards such as the Geforce GTX 1080 Ti.

\subsubsection{NVIDIA workstation graphics cards}

A dataset gathering information on 167 NVIDIA workstation graphics cards models released between 2013 and 2023 included is curated. The main information gathered for each model includes: Release date, die area, technological node, memory type, memory size, \textit{Thermal Design Power} (TDP) and compute power (Single, Double, Tensor and Half floating precision).
The dataset is based on data retrieved from the TechPowerUp GPU database\footnote{ \url{https://www.techpowerup.com/gpu-specs/} on 12/12/2023}. 
Another dataset of 76 graphics cards models based on a wikipedia page listing NVIDIA graphics cards\footnote{\url{https://en.wikipedia.org/wiki/List_of_Nvidia_graphics_processing_units} on April 12, 2024} was retrieved.
We merged the two datasets to cross-validate the specifications of the different cards. This validated information on 74 out of the 167 models ($44\%$)
In cases of divergent information, NVIDIA's published datasheets are taken as reference. 
Our final dataset is available with the accompanying code.   

\subsubsection{Other graphics cards}

To assess the environmental impacts of \gls{ml} models, we gather information on other graphics card models used for training.
We relied on different sources: the TechPowerUp database provided details for NVIDIA non-workstation cards (11 card models), the Google Cloud Platform documentation and publications by Google on their hardware, and website of the manufacturer/press releases/benchmarks for the other cards (Cerebras CS-2, Huawei Ascend 910 and AMD Instinct MI250X).

\subsection{\gls{ml} models training impacts}
\label{subsec:Machine-learning}

The studies on \gls{ml} models training have been conducted using the Epoch AI Notable systems database \cite{epoch2022pcdtrends} retrieved on July 19, 2024. This database is the most comprehensive one on \gls{ml} systems to our knowledge. It gathers extensive information on a large variety of \emph{notable} \gls{ml} systems (including but not limited to: name, publication date, number of \emph{floating point operations} (FLOP) during training, origin country of model producers, training duration, training hardware and hardware quantity). Notable systems are "Models that have advanced the state of the art, had a large influence in the field’s history, or had a large impact within the world"\footnote{\url{https://epochai.org/data/notable-ai-models-documentation}}. To estimate the environmental impacts of training a \gls{ml} model, information on the training duration, training hardware, hardware usage during training and source of electricity is needed \cite{Morand2024MLCA}.

\begin{table*}[ht]
    \centering
    \begin{tabular}{c|c|c|c c c|c c c c}
        \multicolumn{2}{c|}{} & Systems & FLOP & hardware & both & \thead{training\\duration} & \thead{hardware\\quantity} & both & + hardware \\
        \hline
        \multicolumn{2}{c|}{Number} & 825 & 386 & 244 & 206 & 142 & 166 & 112 & 107 \\
        \multicolumn{2}{c|}{Coverage (\%)} & 100 & 46.8 & 29.6 & 24.9 & 17.2 & 20.1 & 13.6 & 13.0\\
        \hline
        \multirow{4}{*}{Confidence} & Confident & 170 & 116 & 109 & 97 & 77 & 72 & 59 & 58\\
        & Likely & 96 & 65 & 54 & 46 & 36 & 30 & 20 & 18\\
        & Speculative & 51 & 36 & 13 & 13 & 20 & 10 & 9 & 7\\
        & Unknown & 508 & 169 & 63 & 50 & 33 & 30 & 24 & 24
    \end{tabular}
    \caption{Description of the Epoch AI dataset. The values show the number of entries for each information type. Confidence scores are reported by database authors for Training compute, Parameters, and Training dataset size.}
    \label{tab:dataset_specs}
\end{table*}

\subsubsection{Ambiguous card names}
Multiple card names are documented for 5 \gls{ml} models (one per training process in multiple steps carried on different hardware). These models are excluded from analyses as it is unclear which hardware is used for which share of the processes. For some other models the hardware name itself is ambiguous. For instance, it might be indicated as "A100", which could refer to several cards.
Each of these graphics cards can differ in some characteristics (e.g., estimated environmental impact, energy consumption or compute power). In such cases, the most plausible candidate is chosen as the reference value, and the other options are used to compute a value interval.

\subsubsection{Training duration}

\newcommand{\firstestimate}[0]{\texttt{GPU-h\textsubscript{1}}}
\newcommand{\secondestimate}[0]{\texttt{GPU-h\textsubscript{2}}}

The number of GPU-hours required for training a model can be estimated in two ways.
For models specifying duration and hardware quantity, these two values can be multiplied to obtain an estimate referred to as  \emph{\firstestimate{}}. It is the most reliable as it uses information retrieved directly from the papers presenting the models.
However, Table~\ref{tab:dataset_specs} shows the low coverage of this method in the database.
For models were the compute required for training (FLOP) and hardware used are specified, another option is to divide the number of FLOP by the compute power of the card used.
The compute power values that are used are the maximum of the peak performance in single, half  or tensor floating point precision. This second estimation, referred to as \emph{\secondestimate{}}, should lead to an under-estimation of the number of GPU-hours, as the hardware does not always operate at peak performance.
For each of the 106 models where both estimates are available, we compare them to obtain an estimate for the ratio of hardware performance and to validate the estimates from \secondestimate{}.
this process revealed anomalies, especially on fine-tuned models where  \secondestimate{} corresponds to training the base model and \firstestimate{} correspond to the fine-tuning process.
Anomalies also occur when incorrect compute power values are used to estimate the number of FLOP in the EpochAI data-base. These anomalies should not pose problems as we use \firstestimate{} for these models. 

We build a linear model to predict \firstestimate{} using \secondestimate{}. This model is computed on 87 observation excluding the anomalies. 
To ensure a linear relation, we estimate $\log(\firstestimate{}) \sim \log(\secondestimate{})$
The linear regression analysis revealed a statistically significant model (F(1,85) = 4161, p $< 2.2e-16$), with an adjusted R² of 0.98, meaning that 98 percent of the variance in the observations is explained by our model. 
The model equation is $\log(\firstestimate{}) = 1.31 + 1.00 \log(\secondestimate{})$ with a standard error of $0.16$ for the intercept and $0.02$ for the regression coefficient. This indicates that an increase of 1 for the $\log(\secondestimate{})$ value leads to an average increase of 1.00 units in $\log(\firstestimate{})$. This positive relationship between $\log(\secondestimate{})$ and $\log(\firstestimate{})$ was found to be statistically significant (t(85) = 64.50, p $< 2e-16$), affirming the predictive power of $\log(\secondestimate{})$ on $\log(\firstestimate{})$.
In addition to the regression analysis, a scatterplot with the fitted regression line were examined to ensure model assumptions were met. The residuals were normally distributed (Shapiro-Wilk W = .98, p = .24), homoscedasticity was confirmed (studentized Breusch-Pagan test = .86 , p = .35), and the residuals appeared to be independent (Durbin-Watson D = 1.67, p = .055).
This model correspond to using a constant performance ratio of $\simeq 27\%$
The final estimation for the number of GPU hours is thus as follows: we use \firstestimate{} for the 112 models where it is available and use \secondestimate{} for another 93 models. 

\subsubsection{Server characteristics}

For NVIDIA workstations, we supposed that servers contain 4 graphics cards and 2 CPUs. For NVIDIA non-workstation cards, we 
supposed servers with 2 graphics cards for 2 CPUs. Without information on memory usage per model training, we did not account for the presence of memory in the servers, which largely underestimates the production impacts of the servers.
For non-NVIDIA hardware, we searched documentation (e.g., Google Cloud Platform documentation, publications by Google) to obtain information on the number of chips and processors per server. For instance, for TPUv3, a server with two CPUs manages every four TPU chips.

We use values consistent with hyper-scale datacenters for the hardware lifespan, average utilization over its life-cycle and infrastructure energy consumption. We use a lifespan of 3 years for the hardware \cite{Ostrouchov2020gpu}, and, using information from Meta, we choose a close to optimal PUE of 1.1 and a hardware utilization of 50\
Using a constant PUE of 1.1 overlooks the significant reduction in PUE of datacenters used to train \gls{ml} models between 2012 and 2018 \cite{Masanet2020recalibrating}, leading to an under-estimation of the infrastructure consumption for models released before the transition towards hyper-scaler clouds. 

\subsubsection{Hardware consumption}

For the hardware usage during training, a value of 100\% was considered both for GPUs and CPUs. This leads to an overestimate of the energy consumption as processors do not function at max power during the entirety of training. This overestimation should be at most twice the actual consumption value. As we do not account for the memory quantity because of lack of data, we also do not account for its associated energy consumption, leading to an under-estimation of the total energy consumption. This under-estimation should not be too large as processors consumption are generally the dominating source of energy consumption when training \gls{ml} models \cite{Jay2023experimental}.

\subsubsection{Environmental impact of the energy usage}

The energy mix used to evaluate the impacts of each model corresponds to
that 
of the countries 
of the \gls{ml} systems producers. If multiple countries participate in the creation of the system, the energy mixes of each implicated country are considered to create a value interval, and the mix of the country indicated in first position is considered as the reference value.

\subsubsection{Impact of strategies for reducing the environmental impacts of the energy usage}
\label{subsec:carbon_reduction}

Different scenarios of reduction of the carbon intensity of the energy mixes over time are considered.
These scenarios assess the impact of shifting compute locations and investing in decarbonizing electricity for datacenters. The two techniques of compute location shifting and 'greening' the used electricity mixes are important in the strategies of decarbonation of the tech companies \cite{Patterson2022plateau}.
Each scenario comes in the form of a continuous reduction of X\
2019 is chosen as a starting year as it is the year of publication of the paper by \cite{Strubell2019energy} which first raised concerns on the carbon footprint of training \gls{ml} models. 
Scenarios of a reduction  up to $25\%$ per year are explored.
The carbon intensity of the mix used for training a model is thus multiplied by
$(1-\texttt{ratio})^{\texttt{n}}$ where n stand for the number of years since 2019 at release date of the system.

\subsection{Metrics used in this study}

We assess the energy consumption, carbon footprint and metallic resource usage associated with hardware production and training \gls{ml} systems. 
The carbon footprint is assessed according to \emph{Global Warming Potential} (GWP)\footnote{using a time horizon of 100 years (GWP100)},  measured in kg\COtwo{} for the emissions of greenhouse gases such as CO\textsubscript{2} \cite{IPCC2023earth}. Metallic resource usage is assessed through \emph{Abiotic Depletion Potential} (ADPe) measured in \emph{kilograms antimony equivalent} (kg\Sbe{}) \cite{van2020abiotic}. 
Metrics are computed using the MLCA tool \cite{Morand2024MLCA}.

\subsection{Software used in the analysis}

Data manipulation and statistical analyses have been performed using emacs Org mode 9.1.9, Python 3.8.10 using the pandas version 2.0.3 library and R version 3.6.3 (2020-02-29) with the ggplot2\_3.4.3,  dplyr\_1.1.3, lmtest\_0.9-40, stringr\_1.5.0 and zoo\_1.8-12  libraries.  Platform: x86\_64-pc-linux-gnu (64-bit) Running under: Ubuntu 20.04.6 LTS

\section{Impact shifting and hardware production}
\label{sec:impact-shifting}

Computing facilities have been getting increasingly energy efficient and can perform much more computation in a fixed duration using less energy. This efficiency is partly obtained by regular hardware updates in data centers, thus incurring additional equipment production and end-of-life environmental costs.
Frequently changing hardware relies on \textit{impact shifting}, i.e., reducing one type of environmental impact in one specific phase of the life-cycle of a product at the detriment of other categories of environmental impacts or other life cycle phases. Furthermore, graphics cards are getting increasingly more technologically advanced, leading to possible increases in their production impacts.

\subsection{Evolution of graphics cards characteristics}

\begin{figure}[ht]
    \centering
    \includegraphics[width=0.9\columnwidth]{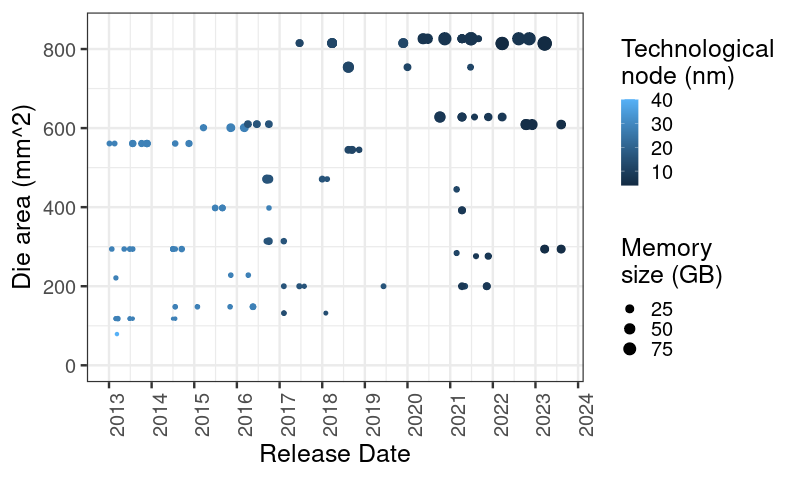}
    \caption{Evolution of the characteristics of NVIDIA workstation graphics cards from 2013 to 2023}
    \label{fig:evol_spec}
\end{figure}

Figure~\ref{fig:evol_spec} shows the evolution of the characteristics of NVIDIA workstation graphics cards from 2013 to 2023. 
Cards are being produced with 
decreasing technological nodes. 
Die area has increased linearly while memory size has grown exponentially 
(around 30\% Cummulative Annual Growth Rate). Exponential growth in the memory size does not necessarily mean a proportional increase in memory IC area as Moore's Law has allowed to exponentially miniaturize ICs.

\begin{figure}[ht]
    \centering
    \includegraphics[width=0.9\columnwidth]{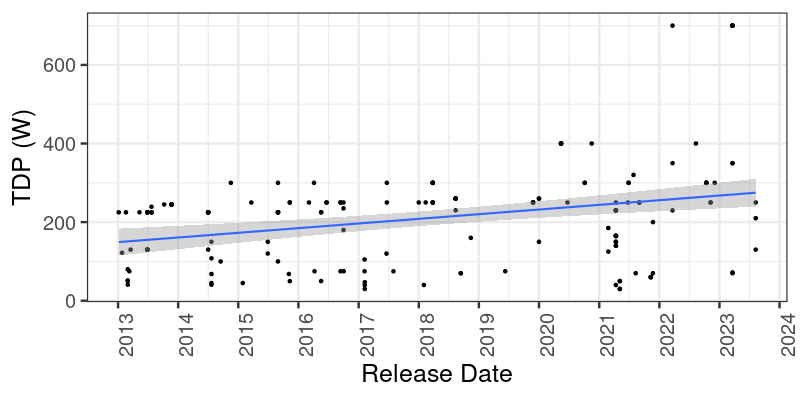}
    \caption{Evolution of the energy consumption of NVIDIA workstation graphics cards from 2013 to 2023}
    \label{fig:evol_TDP}
\end{figure}

Figure~\ref{fig:evol_TDP} shows the evolution of the energy consumption of NVIDIA workstation graphics cards from 2013 to 2023, estimated by their \textit{Termal Design Power (TDP)}. Even if the energy consumption per operation has decreased over time, the total energy consumption of a card has slightly increased over time. This observation indicates a form of rebound effect, where the energy efficiency improvements on the cards have allowed to increase to number of operations realised on a card at a fixed energy consumption.

\subsection{Impacts of graphics cards manufacturing}

\begin{figure*}[t]
\begin{subfigure}{.45\linewidth}
    \centering
    \includegraphics[width=0.9\columnwidth]{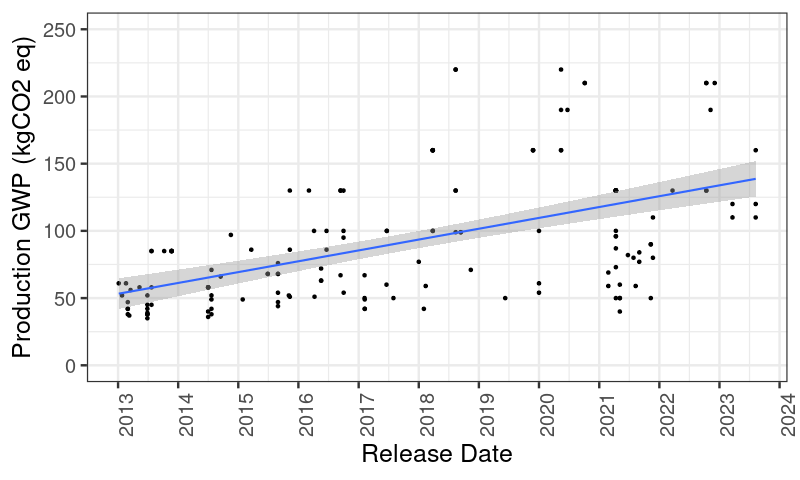}
    \subcaption{GWP}
    \label{fig:evol_gpu_GWP}
\end{subfigure}
\hfill
\begin{subfigure}{.45\linewidth}
    \centering
    \includegraphics[width=0.9\columnwidth]{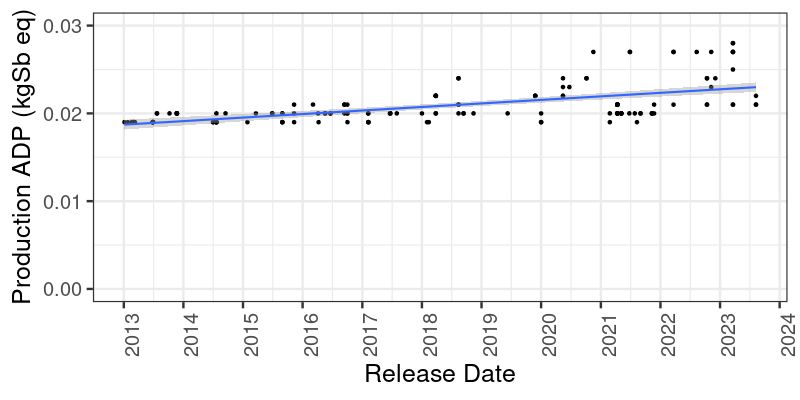}
    \subcaption{ADP}
    \label{fig:evol_gpu_ADP}
\end{subfigure}
    \caption{Evolution of the production impacts (left in GWP, right in ADP) of NVIDIA workstation graphics cards.}
     \label{fig:evol_gpu}
\end{figure*}

 Figure \ref{fig:evol_gpu} shows the evolution in the estimated environmental impacts of graphics cards production, in terms of (GWP) and (ADP). 
 The overall trend suggests that the production impacts have increased over time for both impact categories, as was expected from the observations on the graphics cards characteristics.
Overall, all the important characteristics of graphics cards produced by NVIDIA for workstation have increased, 
leading to an increase in the environmental impacts incurred by the production of these cards over time.

\subsection{Evolution of the hardware requirements}

\begin{figure*}
\begin{subfigure}{.45\linewidth}
        \centering
    \includegraphics[width=0.95\columnwidth]{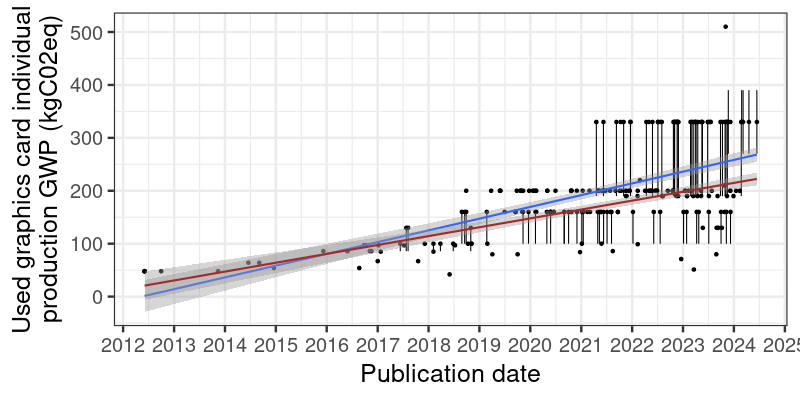}
    \subcaption{GWP}
    \label{fig:production_GWP_used_GPU}
\end{subfigure}
\hfill
\begin{subfigure}{.45\linewidth}
    \centering
    \includegraphics[width=0.95\columnwidth]{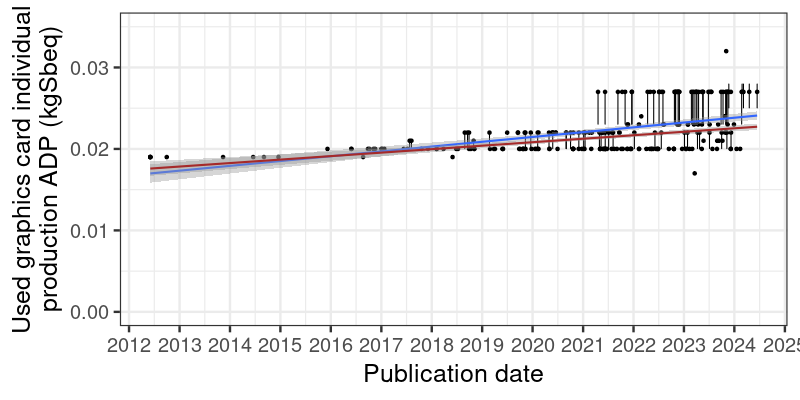}
    \caption{ADP}
    \label{fig:production_ADP_used_gpu}
\end{subfigure}
    \caption{Evolution of the production impacts (left in GWP, right in ADP) of the graphics cards used for training \gls{ml} systems in the Epoch AI data-set. Value intervals correspond to cases of ambiguous card names. The blue line represents the trend in the reference values while the red line represents the trend on the minimal values.}
    \label{fig:production_used_gpu}
\end{figure*}

Figures~\ref{fig:production_GWP_used_GPU} and \ref{fig:production_ADP_used_gpu} show the production impacts of graphics cards used to train \gls{ml} systems in the Epoch AI data-set. The models of graphics cards used to train \gls{ml} systems have followed a similar trend to that of the produced workstation cards, confirming that the production impacts of graphics cards used to train \gls{ml} models have increased over time.

\begin{figure}[t]
    \centering
    \includegraphics[width=0.95\columnwidth]{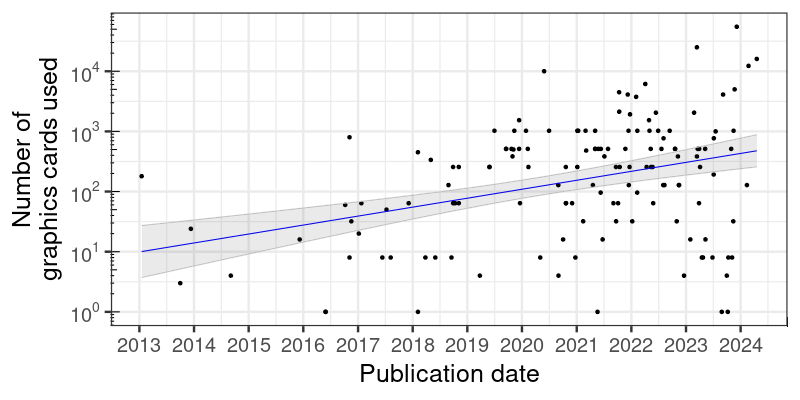}
    \caption{Evolution of the number of graphics cards used for training \gls{ml} models in the Epoch AI dataset. Trend was computed using the \emph{Weighted Least Square} (WLS) method to account for heteroscedasticity}
    \label{fig:hardware_quantity}
\end{figure}

Figure~\ref{fig:hardware_quantity} shows the evolution of the number of graphics cards used to train models in the Epoch AI database. Hardware quantity has increased exponentially. Combining this observation with the observation that the graphics cards used are more impacting at production indicate that the production of the hardware used to train \gls{ml} models cause significant and growing environmental impacts that need to be addressed as well as the energy consumption of the hardware.    

\section{Optimization Strategies in a growing sector}

\label{sec:rebound}

\subsection{Trends in \gls{ml} models training}

\subsubsection{Energy consumption of models training}

\begin{figure}[ht]
    \centering
    \includegraphics[width=.95\columnwidth]{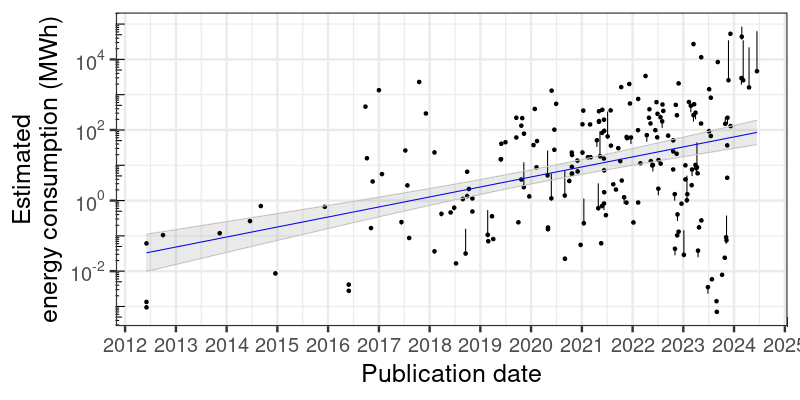}
    \caption{Evolution of the energy consumption of training \gls{ml} models over time. Value intervals account for ambiguous card names. Trend  has been computed using WLS to account for heteroscedasticity.}
    \label{fig:dynamic_energy_consumption}
\end{figure}

Figure~\ref{fig:dynamic_energy_consumption} 
shows that energy consumption has increased exponentially 
even when low-energy demanding models are used.
There are too few low energy intensive models to draw reliable conclusions.
Nonetheless, data suggests an increase from 2012 to 2019, and a decrease from 2019 to 2023, bringing energy consumption to the level of 2012. 
The advent
of green AI research from 2019
could be a contributing factor. 
The less energy intensive models could also be fine-tuned versions of pre-existing models that build on the larger models to attain good performance with a small overhead.

\subsubsection{Environmental impacts of model training}

\begin{figure*}[t]
\begin{subfigure}{.45\linewidth}
    \centering
    \includegraphics[width=.95\columnwidth]{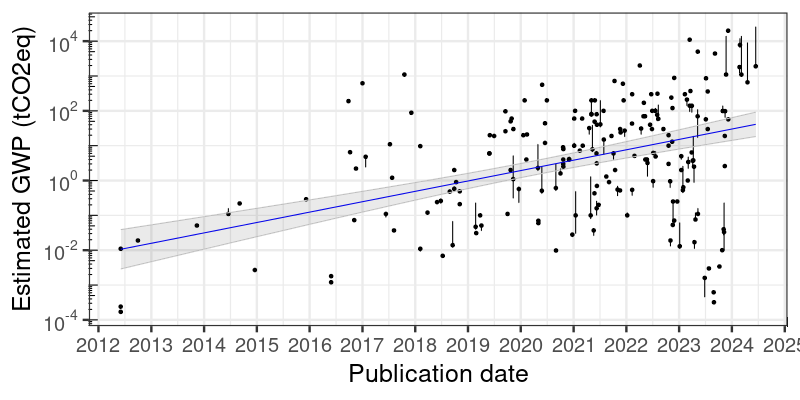}
    \subcaption{GWP}
    \label{fig:GWP_training}
\end{subfigure}
\hfill
\begin{subfigure}{.45\linewidth}
    \centering
    \includegraphics[width=.95\columnwidth]{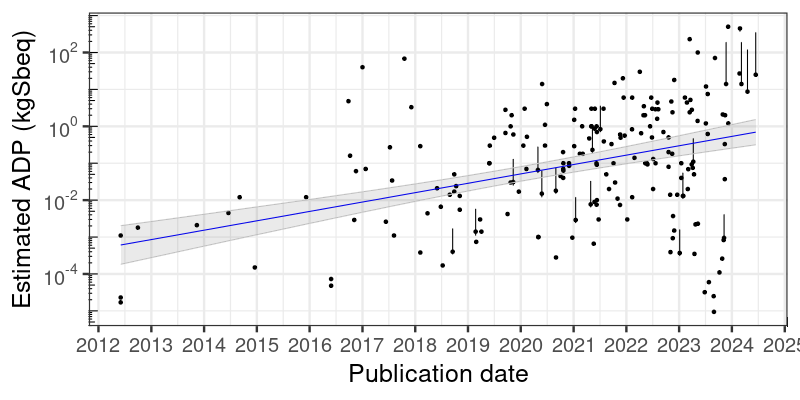}
    \subcaption{ADP}
    \label{fig:ADP_training}
\end{subfigure}
 \caption{Environmental impacts in GWP (a) and ADP (b) of training \gls{ml} models over time. Value intervals account for ambiguous card names and model producers from multiple countries. Trends computed using WLS to account for heteroscedasticity.}
 \label{fig:footprint_training}
\end{figure*}

Figure~\ref{fig:footprint_training} presents the estimated training impacts in terms of GWP (Figure~\ref{fig:GWP_training}) and ADP (Figure~\ref{fig:ADP_training}). Both environmental indicators have increased exponentially between 2012 and 2024. 

Table~\ref{tab:share_adp} summarizes the distribution of the shares of embodied impacts, i.e., the share of the total impacts that can be attributed to hardware production. The estimations for the share of GWP are lower than expected (embodied impacts typically amount to a third of the total footprint of model training \cite{Gupta2020chasing,wu2024efficiency}). This underestimation comes from the absence of accounting for the memory used to train models. ADP, however, comes close to exclusively from hardware production. Solely reducing the energy impacts from \gls{ml} models training will not be sufficient to solve all of the environmental impacts of AI.

\begin{table}[ht]
    \centering
    \begin{tabular}{c|c|c|c|c|c|c}
               & Min   &   Q1 &     Q2   &     Mean  &    Q3   &     Max \\
                \hline
            ADP (\%) & 50  & 100 & 100 &  99.2 &  100 & 100 \\
            GWP (\%) & 5.8 & 11 & 15 & 15 & 16 & 47
    \end{tabular}
    
    \caption{Summary of the share of embodied impacts on the total impacts associated with training models Q1, Q2 and Q3 respectively refer to the first, second and third quartiles.}
    \label{tab:share_adp}
\end{table}

\subsection{Is location shifting the solution?}

\begin{figure}[ht]
    \centering
    \includegraphics[width=.95\columnwidth]{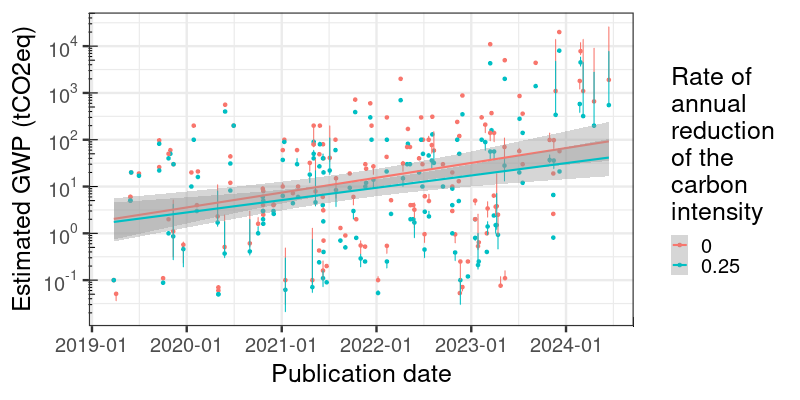}
    
    \caption{Estimated carbon footprint of training \gls{ml} models released after 2019, with or without reduction of the carbon intensity of the used electricity.}
    \label{fig:reduction_CI}
\end{figure}

We compare the impact of models trained with real electricity mixes with a simulated annual reduction of 25\% of the carbon intensity of the electricity mixes. Models with a carbon footprint inferior to 50 kg\COtwo{} are excluded (16 models in real electricity mixes, vs. 21 in the simulation). Figure~\ref{fig:reduction_CI} shows that the
carbon footprint of models released from 2019 is increasing, suggesting that the reduction
strategy 
fails to curb the exponential growth of models impact.

It can be noted that the carbon intensity of electricity mixes is bound to around 15-20g\COtwo{}/kWh, based on the current world lower intensity mixes. 
Thus, even if carbon intensity reductions were quick enough to counterbalance the increase in energy consumption, 
the benefits
would be short lived, under a steady trend in energy consumption. In addition, the production phase of hardware also has an increasing impact which is not 
curbed by reducing carbon intensity.
Finally, the geographic changes intended to reduce carbon intensity could incentivizes 
higher impact for datacenter facilities \cite{Velkovka2015impermanent}.

\label{sec:Limitations-Discussion}
\section{Discussion}

\noindent\textbf{Rebound effect is prevalent in the AI sector} Our observations on the trends in \gls{ml} models training energy consumption, showing the prevalence of the rebound effect are consistent with diverse macro observations: 
\citet{Wu2022sustainable} identify a rebound effect at Facebook while~\citet{Patterson2022plateau} show that a constant share of Google's energy consumption is attributable to AI when the total consumption increased dramatically.
The conclusions of our study should also apply to inferences. \citet{devries2023growing} discusses rising inferences and \citet{Desislavov2023trends} suggest that inferences are also getting 
more computationally and energy expensive over time.

\noindent\textbf{Greener energy cannot void carbon impact.}
As shown in Figure~\ref{fig:reduction_CI}, reduction of the carbon intensity of the electricity used seems insufficient to curb the exponential growth of the carbon footprint of training \gls{ml} models. Furthermore, the electricity consumption of data-centers destabilizes local electricity grids \cite{Ortar2023powering}, potentially causing the prolongation of fossil fuel power plants \cite{FT2024coal, Guardian2024Ireland}. Matching the carbon footprint of AI through carbon offsetting also has limited potential~\cite{Guardian2023carbon,Lohmann2009toward,Bol2021Moore}.

\noindent\textbf{Impacts go beyond carbon footprint}
Our study focused on the carbon footprint and metallic resource depletion incurred by training models. Both metrics have increased over time, but the carbon footprint comes mostly from the energy consumption while metallic resource depletion comes from hardware production, showing the importance of multi-criteria environmental assessments.
AI also causes multiple other environmental impacts that are harder to quantify, 
such as water usage~\cite{Mytton2021datacenter}, 
that creates a competition for resources, especially in urban areas \cite{Roussilhe2024silicon}. 

The mining process and the disposal of hardware at its end of life also cause the destruction of ecosystems \cite{Comber2023Computing} and create multiple pollutants that are toxic to human and non-human life \cite{WHO2021children}.
AI also has significant social consequences and poses numerous ethical challenges \cite{Bender2021parrots,Jiang2023Art,Dauvergne2021globalization,Keyes2018misgendering, Shi2023geographically}.

\section{Limitations}

We acknowledge limitations in our study, which stem from the material and methods available to conduct the analysis. Nonetheless, this remains, to our knowledge, the most comprehensive study on the global impact of the AI sector.

\noindent \textbf{Limitations of the material and methods.} 
The production impact assessment in MLCA does not account for technological node, and supposes a fixed memory density (while it has increase over time), leading to production impacts estimate being inexact.
MLCA currently does not account for the end of life of the hardware nor for the life cycle of datacenter buildings or cooling infrastructure. 
The Epoch AI database is incomplete (models and model information are missing), which leads to some uncertainty on the observed trends, as the excluded models could have an impact.

\noindent \textbf{Modeling hypotheses are needed for analysis.}
Hypotheses sometimes lead to under or over-estimation of some parameters such as GPU consumption or training duration.
For example, the impact of the BLOOM model is overestimated in our study. It listed in Epoch AI as an international collaboration, so the world average electricity mix is used in our analysis although it was trained in France, which has a lower carbon intensity than world average. 
Lacking information, we supposed that all models were trained in hyper-scale data-centers, leading to using a quasi optimal PUE for all models, masking the increase in PUE over the last decade.

\noindent \textbf{This study is limited to model training. } It does not account for inferences (which are becoming increasingly prevalent over the life cycle of models),  model retraining or multiple training iterations during model development.

\section{Conclusion and Future work}
\label{sec:conclusion}

In this paper, we curated a graphics cards dataset and showed that the production impacts of the hardware used for training \gls{ml} models have continuously increased over time. 
Using the Epoch AI dataset, we 
showed that the energy consumption and environmental impacts associated with training models have also increased, even when considering reduction strategies such as location shifting. 
We demonstrate the limits of impact shifting strategies. Our results suggest that current impact reduction strategies alone cannot curb the growth in the environmental impacts of AI. 
Impact reduction must be combined with a broader reflection on the place and role of AI in a sustainable society.

\bibliography{biblio.bib}

\end{document}